%% file: main.tex
\documentclass{article}

\usepackage[preprint, nonatbib]{neurips_2024}

\usepackage[utf8]{inputenc} 
\usepackage[T1]{fontenc}    
\usepackage{hyperref}       
\usepackage{url}            
\usepackage{booktabs}       
\usepackage{amsfonts}       
\usepackage{nicefrac}       
\usepackage{microtype}      
\usepackage{wrapfig}

\usepackage{subfigure}
\usepackage[ruled]{algorithm2e}
\usepackage{bbm}
\usepackage{amsfonts}
\usepackage{epsfig}
\usepackage{graphicx}
\usepackage{amsmath,bm}
\usepackage{amssymb}
\usepackage{bbding}
\usepackage{color}
\usepackage{xcolor}
\usepackage{cleveref}

\usepackage{wrapfig}

\usepackage{lipsum}
\newcommand\blfootnote[1]{%
  \begingroup
  \renewcommand\thefootnote{}\footnote{#1}%
  \addtocounter{footnote}{-1}%
  \endgroup
}

\crefname{figure}{figure}{figures}
\Crefname{figure}{Figure}{Figures}
\hypersetup{colorlinks,linkcolor={red},citecolor={green},urlcolor={blue}}

\hypersetup{
    colorlinks=true,
    urlcolor=magenta,
}

\usepackage{etoolbox} 
\newcommand{\MyMapTemplatePrefixc}[4]{\expandafter#1\csname#3#4\endcsname{#2{#4}}} 
\forcsvlist{\MyMapTemplatePrefixc {\def} {\mathcal}{c}} {A,B,C,D,E,F,G,H,I,J,K,L,M,N,O,P,Q,R,S,T,U,V,W,X,Y,Z}  

\newcommand{\MyMapTemplatePrefixtb}[5]{\expandafter#1\csname#4#5\endcsname{#2{#3{#5}}}} 
\forcsvlist{\MyMapTemplatePrefixtb {\def} {\tilde}{\mathbf}{t}} {A,B,C,D,E,F,G,H,I,J,K,L,M,N,O,P,Q,R,S,T,U,V,W,X,Y,Z}  
\forcsvlist{\MyMapTemplatePrefixtb {\def} {\tilde}{\mathbf}{t}} {0,1,a,b,c,d,e,f,g,h,i,j,k,l,m,n,o,p,q,r,s,u,v,w,x,y,z}  

\newcommand{\MyMapTemplateNoPrefix}[3]{\expandafter#1\csname#3\endcsname{#2{#3}}}
\forcsvlist{\MyMapTemplateNoPrefix {\def} {\mathbf} } {0,1,a,b,c,d,e, f, g, h, i, j, k, l, m, n, o, p, q, r, u, v, w, x, y, z} 
\forcsvlist{\MyMapTemplateNoPrefix {\def} {\mathbf} } {A,B,C,D,E,F,G,H,I,J,K,L,M,N,O,P,Q,R,S,T,U,V,W,X,Y,Z}  

\definecolor{sotacolor}{RGB}{220, 239, 220}
\definecolor{tobecite}{RGB}{100, 180, 236}

\usepackage{caption}
\usepackage{bbding}

\title{Direct3D: Scalable Image-to-3D Generation via 3D Latent Diffusion Transformer}

\author{%
Shuang Wu$^{1,2}$\thanks{Equal contribution.} \quad Youtian Lin$^{2*}$ \quad  Feihu Zhang$^{1}$ \quad Yifei Zeng$^{1,2}$ \quad Jingxi Xu$^{1}$   \AND
\quad Philip Torr$^{3}$\thanks{Chief scientific advisor of DreamTech, all work was done at DreamTech.}  \quad Xun Cao$^{2}$ 
\quad Yao Yao$^{2}$\thanks{Corresponding author.} \\ \\
$^{1}$DreamTech \qquad $^{2}$Nanjing University \qquad $^{3}$University of Oxford\\
}

\begin{document}

\maketitle

\input{sections/abstract}

\input{sections/introduction}

\input{sections/relatedwork}

\input{sections/method}

\input{sections/experiment}


\input{sections/conclusion}

\appendix


{\small
\bibliographystyle{ieee_fullname}
\bibliography{egbib}
}%

\clearpage
\input{sections/appendix}


\end{document}


\maketitle

\section{Distant Function Selection}
We evaluate the distant function with piece-wise step function $\mathbf{H}(\cdot)$ and sigmoid function $\mathbf{S}(\cdot)$, as shown in Fig.~\ref{fig:h(x)} and Fig.~\ref{fig:s(x)} respectively.
The experimental results based on RetinaNet~\cite{Lin_2017_retina} are given in Table~\ref{table:varying_lambda_delta}. We can observe that the performance gap between piece-wise step function $\mathbf{H}(\cdot)$ and sigmoid function $\mathbf{S}(\cdot)$ is only $0.1\%$ in term of AP~($37.4$ \emph{v.s.} $37.3$).
The results demonstrate that these two distance functions have no essential difference.
In this paper, we use $\lambda = 8$ for all experiments.


\begin{minipage}[t!]{\textwidth}
\begin{minipage}[t!]{0.24\textwidth}
\setlength{\abovecaptionskip}{0.cm}
\setlength{\belowcaptionskip}{0.cm}
\makeatletter\def\@captype{figure}
\centering
\includegraphics[width =\textwidth]{h(x).pdf}
\caption{The shape of $\mathbf{H}(\cdot)$.}
\label{fig:h(x)}
\end{minipage}
\begin{minipage}[t!]{0.24\textwidth}
\setlength{\abovecaptionskip}{0.cm}
\setlength{\belowcaptionskip}{0.cm}
\makeatletter\def\@captype{figure}
 \centering
 \includegraphics[width =\textwidth]{sigmoid.pdf}
\caption{The shape of $\mathbf{S}(\cdot)$.}
\label{fig:s(x)}
\end{minipage}
\begin{minipage}[t!]{0.5\textwidth}
\setlength{\abovecaptionskip}{0.cm}
\setlength{\belowcaptionskip}{0.cm}
\makeatletter\def\@captype{table}
\vspace{-0.5cm}
\caption{Varying $\delta$ and $\lambda$ for distance function.}
\resizebox{\linewidth}{!}{
\begin{tabular}{c|lll||c|lll}
\hline
\hline
\multicolumn{1}{c|}{$\delta$}& \multicolumn{1}{c}{$\mathrm{AP}$}& \multicolumn{1}{c}{$\mathrm{AP}_{50}$}& \multicolumn{1}{c||}{$\mathrm{AP}_{75}$}&
 \multicolumn{1}{c|}{$\lambda$}&\multicolumn{1}{c}{$\mathrm{AP}$} & \multicolumn{1}{c}{$\mathrm{AP}_{50}$} & \multicolumn{1}{c}{$\mathrm{AP}_{75}$}\\
\hline
\hline
1 & 37.0 & 57.6& 39.2& 2& 36.4& 57.1& 37.9\\
0.5 & \textbf{37.4} & 57.5& 39.2& 4& 36.9& 57.5& 38.7\\
0.25 & 36.8& 56.3& 38.7& 8& \textbf{37.3}& 57.4& 38.9\\
0.125 & 35.1& 53.8& 36.6& 16& 36.5& 55.9& 38.3\\
\hline
\hline
\end{tabular}
}
\label{table:varying_lambda_delta}
\end{minipage}
\end{minipage}

\section{The Equivalence between Cross Entropy and Error-Driven Update}
Here we find that if pair-wise error loss has the same gradients form as Eq.(7) in the main paper, then Error-Driven Update can be omitted for simplicity.
To keep the numerator of pair-wise error gradients as the same as Eq.(7) in the main paper, we follow the common practice on cross entropy loss which adds a logistic function to sigmoid function.
To start with, $\mathbf{S}(\cdot)$ is replaced with $\mathbf{CE}(\mathbf{S}(\cdot),0)$, which can be written as:
\begin{equation}
\begin{small}
\begin{aligned}
CE(S(\hat{P}_{v}-\hat{P}_{u}),0) &= -\frac{1}{\lambda }((1-0)\cdot log(1-S(\hat{P}_{v}-\hat{P}_{u}))+0\cdot (S(\hat{P}_{v}-\hat{P}_{u})))\\
&=-\frac{1}{\lambda }log(1-S(\hat{P}_{v}-\hat{P}_{u}))
\end{aligned}
\end{small}
\label{eq.10}
\end{equation}
where the gradients of this distance function w.r.t $S(\hat{P}_{v}-\hat{P}_{u})$ can be calculated as:
\begin{equation}
\begin{small}
\begin{aligned}
\frac{\partial CE(S(\hat{P}_{v}-\hat{P}_{u}),0)}{\partial S(\hat{P}_{v}-\hat{P}_{u})} = \frac{1}{\lambda (1-S(\hat{P}_{v}-\hat{P}_{u}))}
\end{aligned}
\end{small}
\label{eq.11}
\end{equation}
Since the gradient of $S(\hat{P}_{v}-\hat{P}_{u})$ \wrt $\hat{P}_{u}$ can be written as:
\begin{equation}
\small
\label{eq.12}
\frac{\partial S(\hat{P}_{v}-\hat{P}_{u})}{\partial\hat{P}_{u}} = -\lambda S(\hat{P}_{v}-\hat{P}_{u})(1-S(\hat{P}_{v}-\hat{P}_{u}))
\end{equation}
we can have the the gradients of distance function \wrt $\hat{P}_{u}$:
\begin{equation}
\small
\label{eq.13}
\begin{aligned}
&\frac{\partial CE(S(\hat{P}_{v}-\hat{P}_{u}),0)}{\partial \hat{P}_{u}} = \frac{\partial CE(S(\hat{P}_{v}-\hat{P}_{u}),0)}{\partial S(\hat{P}_{v}-\hat{P}_{u})} \cdot \frac{\partial S(\hat{P}_{v}-\hat{P}_{u})}{\partial\hat{P}_{u}}\\
&= \frac{1}{\lambda (1-S(\hat{P}_{v}-\hat{P}_{u}))} \cdot (-\lambda S(\hat{P}_{v}-\hat{P}_{u})(1-S(\hat{P}_{v}-\hat{P}_{u})))\\
&=-S(\hat{P}_{v}-\hat{P}_{u})
\end{aligned}
\end{equation}
Also, to keep the denominator term $BC$ as the same as Eq.~(7) in the main paper, we detach it from backpropagation and treat it as a constant.
Note that, after employing these two tricks (\ie cross entropy and detaching), we can have the same gradient of our pair-wise error (\ie, $(-\sum_{v \in\mathcal{N}}S(\hat{P}_{v}-\hat{P}_{u}))/(rank^+(u)+rank^-(u))$) as AP loss, which theoretically leads to similar performances. The experimental results in Table~1 in the main paper also demonstrate that.

\section{Threshold for Selecting Valid Negative Samples}
In training processing, the number of negative samples $N_{neg}$ is enormous and might overwhelm the loss.
To solve this issue, we utilize a larger margin threshold $T$ to filter out easy negative samples, as shown in Fig.~\ref{fig:easy_pairs}.
Specifically, we set a valid indicator for each pair-wise error to ignore easy pairs. Here we describe indicator function $\mathbbm{1}_{uv}$ as:
\begin{equation}
\begin{small}
\mathbbm{1}_{uv}=\left\{
\begin{aligned}
1,\qquad  & \hat{P}_{v}- \hat{P}_{u}>T \\
0,\qquad  & otherwise
\end{aligned}
\right.
\end{small}
\end{equation}
Then $N_{neg}$ is formulated as: ${N_{neg}} = \sum_{v \in \mathcal{N}}\mathbbm{1}_{uv}$. We also study the impact of different thresholds on detection accuracy. As shown in Table~\ref{table:varying_th_bc}, when $T=0.25$, $N_{neg}$ provides the same performance as $rank^+(u)+rank^-(u)$. This demonstrates the selection of these two balance constants is robust.

\begin{minipage}[t!]{\textwidth}
\vspace{-0.24cm}
\begin{minipage}[t!]{0.5\textwidth}
\setlength{\abovecaptionskip}{-0cm}
\setlength{\belowcaptionskip}{0.cm}
\makeatletter\def\@captype{figure}
\centering
\includegraphics[width=0.75\textwidth]{easy_pairs.pdf}
\label{fig:easy_pairs}
\caption{The comparison with Hard Pair Mining}
\end{minipage}
\begin{minipage}[t!]{0.5\textwidth}
\setlength{\abovecaptionskip}{0.cm}
\setlength{\belowcaptionskip}{0.cm}
\makeatletter\def\@captype{table}
\caption{Varying $th$ for $N_{neg}$.}
\resizebox{\linewidth}{!}{
  \begin{tabular}{c|c|lll}
\toprule[1.5pt]
\multicolumn{1}{c|}{Balance Constant}& \multicolumn{1}{c|}{$T$} & \multicolumn{1}{c}{$\mathrm{AP}$}& \multicolumn{1}{c}{$\mathrm{AP}_{50}$} &\multicolumn{1}{c}{$\mathrm{AP}_{75}$}\\
\hline
\hline
$rank^+(u)+rank^-(u)$& N/A & 37.3& 57.4 & 38.9 \\
\hline
$N_{neg}$ &0 &36.8  & 57.1& 38.8\\
$N_{neg}$ & 0.2 & 37.2 & 57.0& 38.9\\ 
$N_{neg}$ &0.25 &37.3& 56.7 & 39.4 \\
$N_{neg}$ &0.3 &36.9 & 56.3 & 38.7\\
$N_{neg}$ &0.5 & 35.2 &53.3& 37.1\\
\bottomrule[1.5pt]
\end{tabular}
}
\label{table:varying_th_bc}
\end{minipage}
\end{minipage}

\begin{wraptable}{r}{0.4\linewidth}
\vspace{-0.3cm}
\setlength{\abovecaptionskip}{0.cm}
\setlength{\belowcaptionskip}{-0.cm}
    \caption{Varying $Q$ on FCOS~\cite{tian2019fcos}}
\centering
\begin{tabular}{l|lll}
\toprule[1.5pt]
\multicolumn{1}{c|}{$Q$}& 
\multicolumn{1}{c}{$\mathrm{AP}$}& \multicolumn{1}{c}{$\mathrm{AP}_{50}$}& \multicolumn{1}{c}{$\mathrm{AP}_{75}$}\\ 
\hline
\hline
10,000& 37.6& 54.3& 40.0\\
50,000 & 39.7& 57.3& 42.3\\
100,000& 40.0& 58.1& 42.4\\
200,000& 40.0& 58.1& 42.6\\
\bottomrule[1.5pt]
\end{tabular}
\label{tab:M}
\vspace{-0.5cm}
\end{wraptable}
\section{Maximum Pair Number}
In our experiments, the memory (11GB) of \texttt{2080TI} GPU can be ran out because of the extreme large number of pair $\begin{Bmatrix}\hat{P}_{v}, \hat{P}_{u}\end{Bmatrix}$. Thus we adopt a simple yet efficient trick; constricting the input number of pairs.

Here we denote the maximum input number of pairs by $Q$~(\ie the maximum length of $\mathcal{A}_u$ for $L_{\text{APE}}$). Specifically, we manually choose the top $Q$ predictions $\hat{P}_{v}$ of negative samples in $\mathcal{A}_u$. We conduct experiments varying $Q$ for APE loss on FCOS, and the results are shown in Table.~\ref{tab:M}.
When $Q$ is greater than $100,000$, the performance will no longer be improved.
It can be concluded from the results that the promotion from large $Q$ becomes minor as the gradually increasing of $Q$.



{\small
\bibliographystyle{ieee_fullname}
\bibliography{egbib}
}

%% file: sections/abstract.tex
\begin{abstract}

Generating high-quality 3D assets from text and images has long been challenging, primarily due to the absence of scalable 3D representations capable of capturing intricate geometry distributions. In this work, we introduce Direct3D, a native 3D generative model scalable to in-the-wild input images, without requiring a multi-view diffusion model or SDS optimization.
Our approach comprises two primary components: a Direct 3D Variational Auto-Encoder (\textbf{D3D-VAE}) and a Direct 3D Diffusion Transformer (\textbf{D3D-DiT}). D3D-VAE efficiently encodes high-resolution 3D shapes into a compact and continuous latent triplane space. 
Notably, our method directly supervises the decoded geometry using a semi-continuous surface sampling strategy, diverging from previous methods relying on rendered images as supervision signals.
D3D-DiT models the distribution of encoded 3D latents and is specifically designed to fuse positional information from the three feature maps of the triplane latent, enabling a native 3D generative model scalable to large-scale 3D datasets.
Additionally, we introduce an innovative image-to-3D generation pipeline incorporating semantic and pixel-level image conditions, allowing the model to produce 3D shapes consistent with the provided conditional image input.
Extensive experiments demonstrate the superiority of our large-scale pre-trained Direct3D over previous image-to-3D approaches, achieving significantly better generation quality and generalization ability, thus establishing a new state-of-the-art for 3D content creation. Project page: \href{https://nju-3dv.github.io/projects/Direct3D/}{https://nju-3dv.github.io/projects/Direct3D/}.

\blfootnote{This research was supported by DreamTech, and the IP belongs to DreamTech.}


\end{abstract}

%% file: sections/introduction.tex
\section{Introduction}

\begin{figure*}[!t]
  \centering
  \includegraphics[width=1\linewidth]{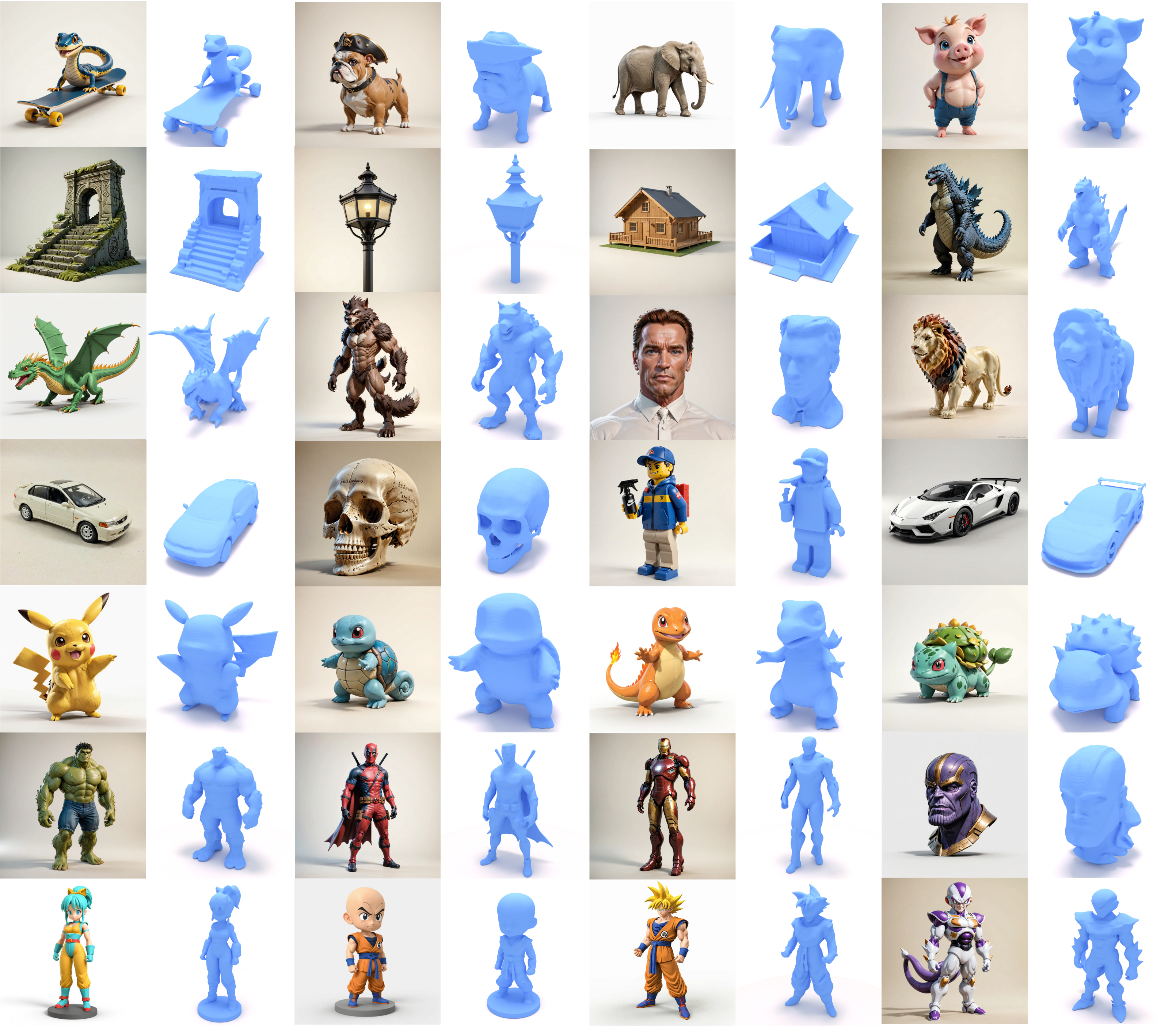}
  \caption{Direct3D is a novel image-to-3D generation method that directly trains on larger-scale 3D datasets and performs state-of-the-art generation quality and generalizability. We achieve this by designing a novel 3D latent diffusion transformer model, which takes an image as the condition prompt and generates high-quality 3D shapes that highly consistent with the conditional images. 
  As shown above, our method can generate 3D shapes from existing text-to-image diffusion models, which indicates that our method generalizes to in-the-wild images, while it only trains on 3D data.
  }
  \label{fig:teaser}
\end{figure*}

In recent years, substantial advancements have been made in 3D shape generation through the utilization of diffusion models~\cite{ho2020ddpm,sohl2015deep}. Inspired by the efficacy demonstrated in text-to-2D image generation, these methods seek to extend the capabilities of diffusion models to the realm of 3D shape generation through extensive training on diverse 3D datasets. Various approaches have explored diverse 3D representations, including point clouds~\cite{mo2024dit3d,nichol2022pointe}, voxels~\cite{ren2023xcube}, and SDFs~\cite{lyu2024getmesh}, aiming not only to capture object appearance faithfully but also to preserve intricate geometric details. However, existing large-scale 3D datasets, such as ObjverseXL~\cite{deitke2024objaversexl}, are constrained both in the quantity and diversity of shapes compared to their 2D counterparts like Laion5B~\cite{schuhmann2022laionb}, which contains 5 billion images, while ObjverseXL only comprises 10 million 3D shapes.

To address this limitation, many existing methods~\cite{chen2023cascade,chen2024microdreamer,gao2024cat3d,liu2023oneplus,liu2023one2345,liu2023zero,liu2023syncdreamer,long2023wonder3d,lu2023direct25,qiu2023richdreamer,zheng2024mvd} employ a pipeline where multi-view images of an object are initially generated from a single image using a multi-view diffusion model. Subsequently, techniques such as sparse view reconstruction methods~\cite{li2023instant3d,long2022sparseneus,wang2023pf,zhang2024gs} or score distillation sampling (SDS) optimization~\cite{poole2022dreamfusion,qiu2023richdreamer,shi2023mvdream,wang2023imagedream} are applied to fuse these multi-view images into 3D shapes. While this pipeline can result in high-quality 3D shape creation, the indirect generation from multi-view images raises efficiency concerns. Additionally, the quality of the resulting shape is heavily dependent on the fidelity of the multi-view images, often leading to detail loss or reconstruction failures.

In this paper, we eschew the conventional approach of indirectly generating multi-view images and instead advocate the direct generation of 3D shapes from single-view images, leveraging a native 3D diffusion model. Inspired by the success of latent diffusion models in 2D image generation, we propose the utilization of a 3D variational auto-encoder (VAE)~\cite{kingma2013vae} to encode 3D shapes into a latent space, followed by a diffusion transformer model (DiT)~\cite{peebles2023dit} to generate 3D shapes from this latent space, conditioned on an image input. 
However, efficiently encoding a 3D shape into a latent space conducive to diffusion model training is challenging, as is decoding the latent representation back into 3D geometry. Previous approaches have employed multi-view images as indirect supervision~\cite{jun2023shapee,lan2024ln3diff,lyu2024getmesh,zhang2024compress3d} through differentiable rendering, but still encounter accuracy and efficiency issues.
To address these challenges, we employ a transformer model to encode high-resolution point clouds into an explicit triplane latent, which has been widely used in 3D reconstruction methods~\cite{chan2022efficient} for its efficiency. While the latent triplane is intentionally set with a low resolution, we introduce a convolutional neural network to upsample the latent resolution and decode it into a high-resolution 3D occupancy grid. Furthermore, to ensure precise supervision of the 3D occupancy grid, we adopt a semi-continuous surface sampling strategy, enabling the sampling and supervision of surface points in both continuous and discrete manners. This approach facilitates the encoding and reconstruction of 3D shapes within a compact and continuous explicit latent space.

For image-to-3D generation, we further leverage an image as the conditional input to the 3D diffusion transformer, which arranges the 3D latent space as a combination of three orthogonal views of a 3D shape. Particularly, we integrate pixel-level image information into each DiT block to ensure the alignment of high-frequency details between the generated 3D models and the conditional images. Furthermore, we introduce cross-attention layers into each DiT block to incorporate semantic-level image information, thereby facilitating the generation of high-quality 3D shapes semantically consistent with the conditional images.


We demonstrate the high-quality 3D generation and strong generalization abilities of the proposed Direct3D approach through extensive experiments. Figure~\ref{fig:teaser} illustrates the 3D generation results of our method on the in-the-wild images generated from Hunyuan-DiT. To summarize, the major contributions of this work include:
\begin{itemize}
    \item We introduce Direct3D, to our best knowledge, the first native 3D generative model scalable to in-the-wild input images. This enables high-fidelity image-to-3D generation without the need for multi-view diffusion models or SDS optimization.
    \item We propose D3D-VAE, a novel 3D variational auto-encoder effectively encoding a 3D point cloud into a triplane latent. Instead of using rendered images as supervision signals, we supervise the decoded geometry directly using a semi-continuous surface sampling strategy to preserve detailed 3D information in the latent triplane. 
    \item We present D3D-DiT, a scalable image-conditioned 3D diffusion transformer capable of generating 3D asserts consistent with input images. The D3D-DiT is specially designed to better fuse the positional information from the latent triplane and effectively integrates pixel-level and semantic-level information from the input image.
    \item We demonstrate through extensive experiments that our large-scale pre-trained Direct3D model surpasses previous image-to-3D approaches in terms of generation quality and generalization ability, setting a new state-of-the-art for the task of 3D content creation.
\end{itemize}

%% file: sections/relatedwork.tex
\section{Related Work}

\subsection{Neural 3D Representations for 3D Generation}
Neural 3D representations are essential for 3D generation tasks. The introduction of Neural Radiance Fields (NeRF) \cite{mildenhall2020nerf}
has significantly advanced 3D generation. Building on NeRF, DreamFusion~\cite{poole2022dreamfusion} introduces a Score Distillation Sampling (SDS) method to generate 3D shapes using an off-the-shelf 2D diffusion model from any text prompt. Many subsequent methods have explored various representations to enhance the speed and quality of 3D generation. For instance, Magic3D \cite{lin2023magic3d} improves generation quality by introducing a second stage using the DMtet~\cite{shen2021deep} representation, which combines Signed Distance Function (SDF) with a tetrahedral grid to represent the 3D shape.


Beyond SDS-based methods, some approaches use directly trained networks to generate different representations~\cite{hong2023lrm,hui2024make,yariv2023mosaic}. For example, LRM~\cite{hong2023lrm} uses triplane NeRF representations as network outputs, significantly speeding up the generation process, albeit with some loss in quality. Another approach, One-2-3-45++~\cite{liu2023oneplus}, proposes using a 3D occupancy grid as the output representation to enhance geometric quality.

\subsection{Multi-view Diffusion}
Following the success of novel view prediction methods using diffusion models, such as Zero123~\cite{liu2023zero}—which generates different unknown views of an object from a single image and text guidance—MVDream~\cite{shi2023mvdream} extends novel view diffusion to generate multiple views of an object at once, improving consistency across views. Imagedream~\cite{wang2023imagedream} further enhances generation quality by introducing a novel image conditional module. Some methods adopt this approach to first generate multi-view images of an object and then reconstruct the 3D shape from these views using sparse reconstruction~\cite{li2023instant3d,long2023wonder3d,wang2023pf,zhang2024gs}. Instant3D~\cite{li2023instant3d} proposes a reconstruction model that takes four multi-view images as input and reconstructs a NeRF representation of the 3D shape. Many subsequent methods have improved on this by enhancing multi-view or reconstruction models~\cite{tang2024lgm, xu2024grm, xu2024instantmesh}.



\subsection{Direct 3D Diffusion}
Despite the challenges of directly training a 3D diffusion model—such as the lack of a diffusible 3D representation—various strategies have been explored. One line of work fits multiple NeRFs to obtain a neural representation of 3D datasets and then applies a diffusion model to generate NeRFs from this learned representation~\cite{shue20233triplanediff}. However, separate training of NeRFs can hinder the diffusion model's ability to generalize to more diverse 3D shapes. 3DGenNeural~\cite{shue20233triplanediff} proposes joint training of triplane fitting of the 3D shape with occupancy as direct supervision to train the triplane reconstruction model.

Another line of work leverages VAEs to encode 3D shapes into a latent space and trains a diffusion model on this latent space to generate 3D shapes~\cite{hong20243dtopia,jun2023shapee,lan2024ln3diff,zhang2024compress3d}. For instance, Shap-E~\cite{jun2023shapee} uses a pure transformer VAE to encode a point cloud and image of a 3D shape into an implicit latent space, which is then recovered into a NeRF and SDF field. 3DGen~\cite{gupta20233dgen} encodes only the point cloud of a 3D shape into an explicit triplane latent space, enhancing generation efficiency. Similar to previous works that fit multiple NeRFs, 3DTopia~\cite{hong20243dtopia} fits multiple triplane NeRFs and encodes the triplane into a latent space for which a diffusion model is trained to generate 3D shapes. Michelangelo~\cite{zhao2024michelangelo} employs 3D occupancy as the output representation for the VAE but uses multiple 1D vectors as implicit latent space instead of a triplane.

However, these methods often rely on rendering loss to supervise the VAE reconstruction, resulting in suboptimal reconstruction and generation quality. Additionally, using implicit latent representations not designed for efficient encoding and lacking compact explicit 3D representations for diffusion further limits their performance. Our D3D-VAE combines the advantages of explicit 3D latent representation and direct 3D supervision to achieve high-quality VAE reconstruction, ensuring robust 3D shape generation. Furthermore, our design for the diffusion architecture specifically addresses conditional 3D latent generation. Our D3D-DiT facilitates pixel-level and semantic-level 3D-specific image conditioning, allowing the diffusion process to generate highly detailed 3D shapes consistent with the condition images.

%% file: sections/method.tex
\section{Methods}

Inspired by LDM~\cite{rombach2022high}, we train a latent diffusion model for 3D generation within a 3D latent space. Unlike previous methods~\cite{jun2023shapee,zhao2024michelangelo} that typically rely on a 1D implicit latent space for generative models, our approach addresses two crucial limitations: 1) the struggle of the implicit latent representation to capture structured information inherent in 3D space, leading to sub-optimal quality of decoded 3D shapes; 2) the challenge of training and sampling from the latent distribution, given that the implicit latent space is unstructured and under-constrained.

\begin{figure*}[!t]
  \centering
  \includegraphics[width=1\linewidth]{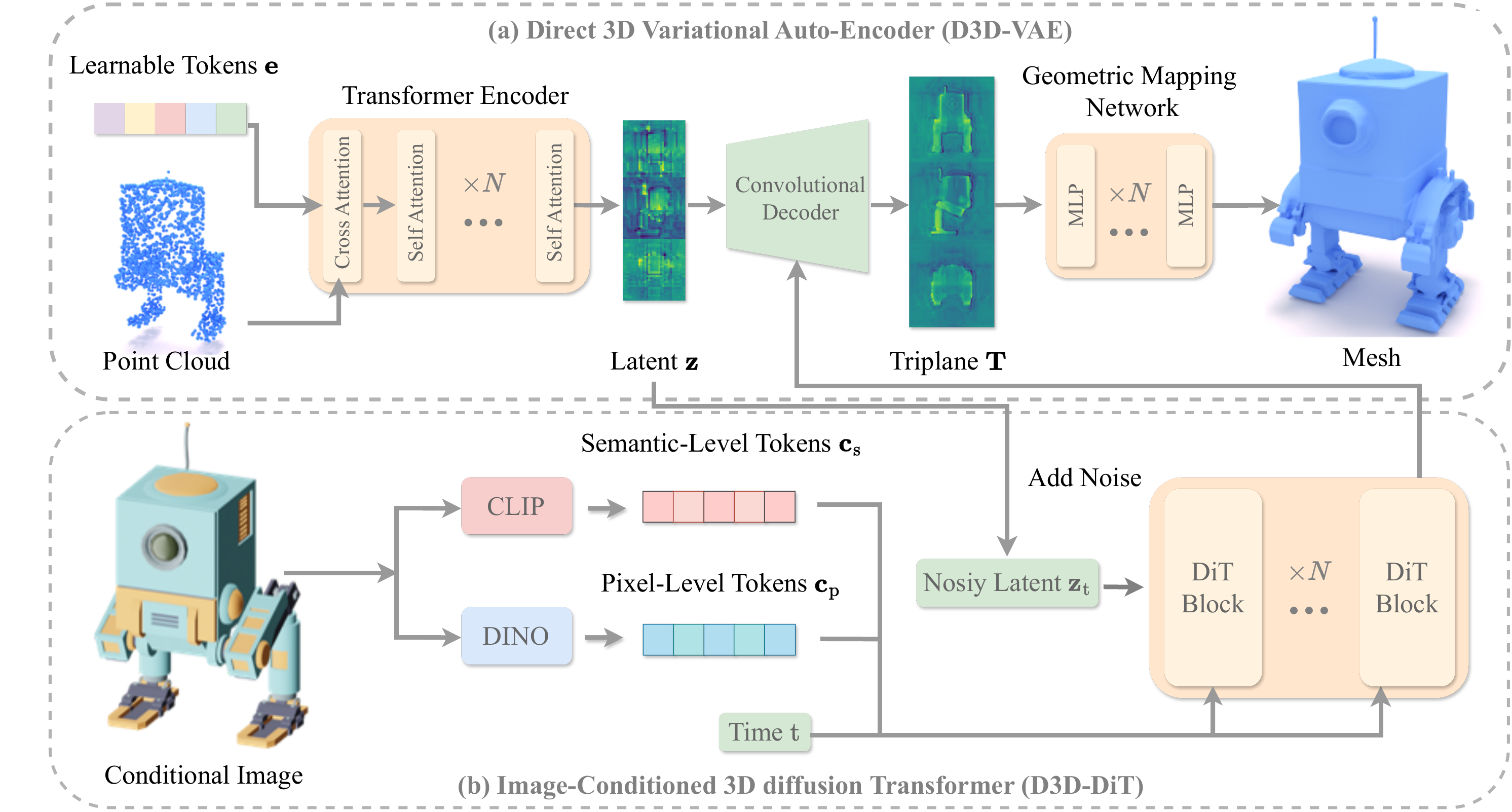}
  \caption{The framework of our Direct3D. (a) We utilize transformer to encode point cloud sampled from 3D model, along with a set of learnable tokens, into an explicit triplane latent space. Subsequently, a CNN-based decoder is employed to upsample these latent representations into high-resolution triplane feature maps. The occupancy values of queried points can be decoded through a geometric mapping network. (b) Then we train the image conditioned latent diffusion transformer in the 3D latent space obtained by VAE. Pixel-level information and semantic-level information from images are extracted using DINO-v2 and CLIP, respectively, and then injected into each DiT block.}
  \label{fig:pipeline}
\end{figure*}

To mitigate these issues, we adopt an explicit triplane latent representation, utilizing a triplane of three feature maps to represent the 3D geometry latent.
The design draws inspiration from LDM, which applies feature maps to represent the 2D image latent. Figure~\ref{fig:pipeline} illustrates the overall framework of our proposed method, which comprises a two-step training process: 1) the D3D-VAE is first trained to convert 3D shapes into 3D latents, which is described in Sec.~\ref{sec:vae}; 2) the image-conditioned D3D-DiT is then trained to generate high-quality 3D assets, which is detailed in Sec.~\ref{sec:dit}.

\subsection{Direct 3D Variational Auto-Encoder}
\label{sec:vae}
The proposed D3D-VAE consists of three components: a point-to-latent encoder, a latent-to-triplane decoder, and a geometry mapping network. Meanwhile, we design a semi-continuous surface sampling strategy that utilizes both continuous and discrete supervision to ensure the high-frequency geometric details of the decoded 3D shape.

\noindent \textbf{Point-to-latent encoder.} In order to obtain robust representations in the latent space that can effectively capture intricate geometry, we uniformly sample high-resolution point clouds from the surface of 3D objects, which is then encoded to an explicit latent representation $\mathbf{z}\in\mathbb{R}^{(3\times r \times r) \times d_\mathbf{z}}$, where $r$ and $\d_\mathbf{z}$ denotes the resolution and channel dimensional of the latent representation, respectively. To be specific, given a set of point clouds $P\in\mathbb{R}^{N_P\times (3+3)}$ sampled from 3D models, where $N_P$ denotes the number of points, the channel dimension $(3+3)$ comprises of the normalized position and normal of each point, we first use Fourier features~
\cite{jaegle2021perceiver} to represent the position structure of point clouds. Then we introduce a series of learnable tokens $\mathbf{e}\in\mathbb{R}^{(3\times r \times r) \times d_\mathbf{e}}$ to query the point cloud features using a cross-attention layer, where $d_\mathbf{e}$ denotes the channel dimensional of $\mathbf{e}$. This enables the injection of 3D information from the point clouds into the latent tokens. Subsequently, multiple self-attention layers are employed to enhance the representation of these tokens, ultimately yielding the latent representation $\mathbf{z}\in\mathbb{R}^{(3\times r \times r) \times d_\mathbf{z}}$, where $d_\mathbf{z}$ represents the channel dimensional of $\mathbf{z}$.

\noindent \textbf{Latent-to-triplane decoder.} After obtaining the latent representation $\mathbf{z}$, we reshape it to the triplane representation. Inspired by RODIN~\cite{wang2023rodin}, we concatenate the three planes vertically along the height dimension, yielding  $\mathbf{z}\in\mathbb{R}^{r\times (3\times r)\times d_\mathbf{z}}$, to prevent incorrect blend of the planes across the channel dimension. Afterwards, the latent-triplane decoder upsamples $\mathbf{z}$ to high-resolution triplane feature maps with upsampling factors $f$. In contrast to the transformer architecture used in the encoder, our decoder model employs convolutional networks to progressively upsample the explicit latent representation and obtain the final triplane $\mathbf{T}=(\mathbf{T_{XY}}, \mathbf{T_{YZ}}, \mathbf{T_{XZ}})$.

\noindent \textbf{Semi-continuous surface sampling.} We employ a Multi-Layer Perceptron (MLP) as the geometric mapping network to predict the occupancy of queried points via features interpolated from the triplane. The MLP contains multiple linear layers with ReLU activation. Typical occupancy is represented by a discrete binary value of 0 and 1 to indicate whether a point is inside an object. However, when the query point is very close to the object surface, it can result in abrupt gradient changes that affect model optimization. In this work, we adopt semi-continuous occupancy, using both continuous and discrete supervision to ensure smooth gradient. Specifically, given a query point $\mathbf{x}$ in 3D space, when its distance to the surface is greater than a small threshold value $s=\frac{1}{512}$, the occupancy value remains either 0 or 1. When the distance is less than s, a continuous value ranging from 0 to 1 is assigned to it. The formula for the semi-continuous occupancy $o(\textbf{x})$ is as follows:
\begin{equation} 
\label{eq:occupancy}
    o(\textbf{x}) = 
        \begin{cases}
            1, & \text{if } sdf(\textbf{x}) < -s \\
            0.5 - \frac{0.5 \cdot sdf(\textbf{x})}{s}, & \text{if } -s \leq sdf(\textbf{x}) \leq s\\
            0, & \text{if } sdf(\textbf{x}) > s
        \end{cases},
\end{equation}
where $sdf(\textbf{x})$ denotes the Signed Distance Function (SDF) value of $\textbf{x}$.

\noindent \textbf{End-to-end optimization.} During the training process, we uniformly sample points from the 3D space and sample points proximate to the object surface to predict their semi-continuous occupancy. We utilize Binary Cross-Entropy (BCE) loss $L_\text{BCE}$ to supervise the predictions. Additionally, we employ KL loss $L_\text{KL}$ to prevent excessive variance in the latent space. Thus, our D3D-VAE 
is optimized by minimizing:
\begin{equation} 
\label{eq:vae_loss}
    L_\text{D3D-VAE}=L_\text{BCE}+\lambda_\text{KL}L_\text{KL},
\end{equation}
where $\lambda_\text{KL}$ denotes the weight of KL regularization.

\subsection{Image-conditioned Direct 3D Diffusion Transformer}
\label{sec:dit}

\begin{wrapfigure}{r}{0.43\textwidth}
  \centering
  \includegraphics[width=0.43\textwidth]{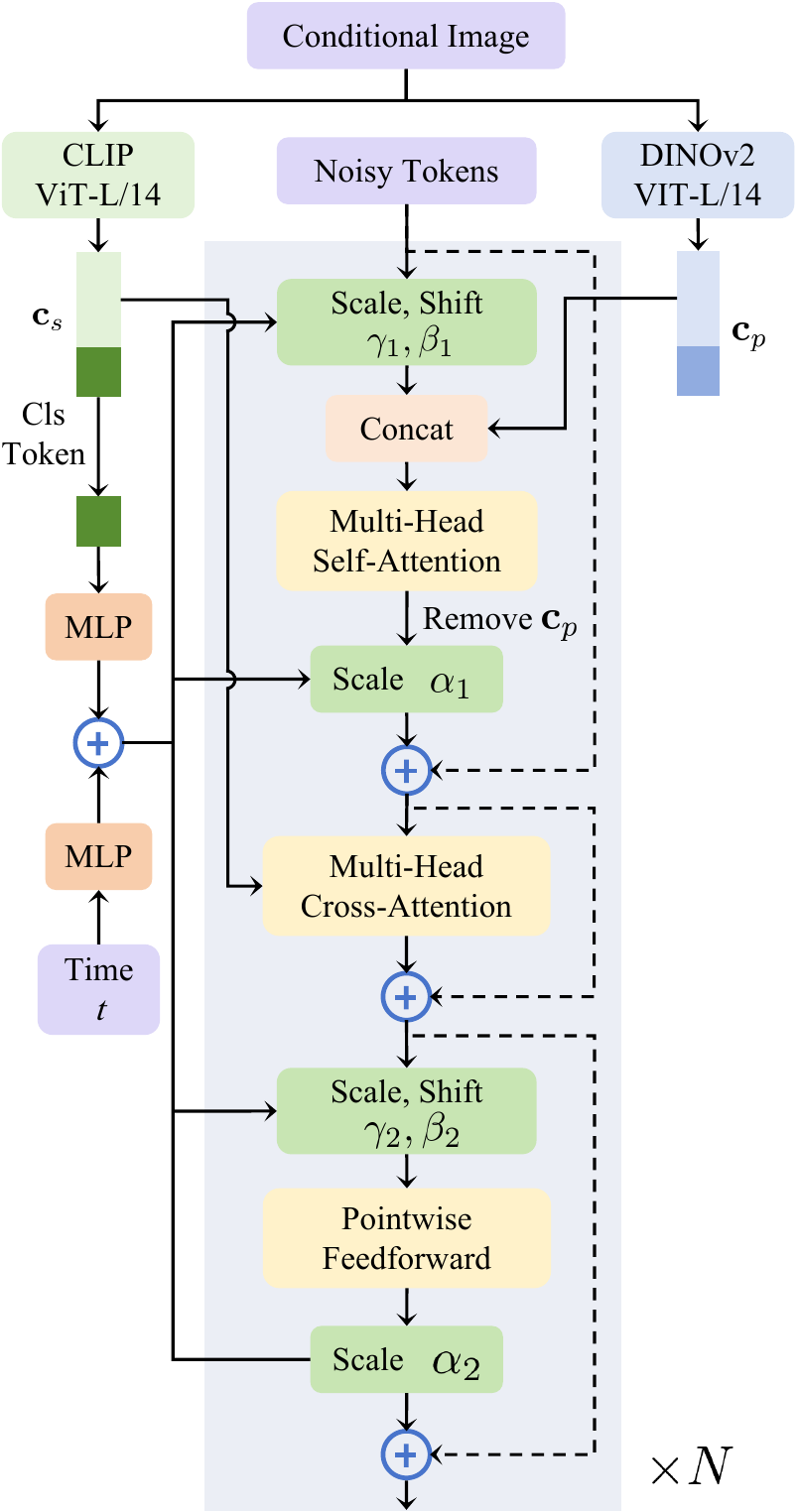}
  \caption{The architecture of our 3D latent diffusion transformer. We employ the pre-trained DINO-v2 and CLIP vision model to extract tokens from conditional images respectively, then incorporate the pixel-level and semantic-level information into each DiT block.}
  \label{fig:dit}
\end{wrapfigure}

After training the D3D-VAE, we have access to a continuous and compact latent space, upon which we train the latent diffusion model. Due to the scarcity of 3D data compared to 2D datasets with billions of images, it is challenging to train a text-conditioned 3D diffusion model with strong generalization. Additionally, text-to-image generative models have already achieved significant maturity, thus we opt to train an image conditional 3D diffusion model for better generalization and higher quality. 

Since the obtained latent embedding is an explicit triplane representation, a naive approach would be to directly use a well-designed 2D U-Net as the diffusion model. However, this would result in a lack of communication between the three planes, thereby failing to capture the structured and intrinsic properties required for 3D generation. Therefore, we build the generation model based on the architecture of the Diffusion Transformer (DiT), utilizing the transformer to better extract spatial positional information among the planes. Meanwhile, we propose to incorporate pixel-level and semantic-level information of the image in each DiT block, thereby aligning the image feature space and latent space to generate 3D assets consistent with the conditional image content. The framework of our latent diffusion model is shown in Figure~\ref{fig:pipeline} \textbf{(b)} and the architecture of each DiT block is illustrated in Figure~\ref{fig:dit}. 

\noindent \textbf{Pixel-level alignment module.} To ensure the high-frequency details of 3D assets generated by the diffusion model are aligned with the conditional images, we design a pixel-level alignment module to inject pixel-level information from the images into the latent space. We employ the pre-trained DINO-v2~\cite{oquab2023dinov2} (ViT-L/14) as the pixel-level image encoder, which has been revealed in previous work~\cite{banani2024probing} to outperform other pre-trained vision models in extracting structural information beneficial for 3D tasks. Specifically, we first use two linear layers with GeLU~\cite{hendrycks2016gelu} activation to project the image tokens $\mathbf{c}_\text{p}$ extracted by DINO-v2 to match the channel dimension of the noisy latent tokens $\mathbf{z}_t$. Then in each DiT block, we concatenate them with the flattened $\mathbf{z}_t$ and feed them into a self-attention layer to model the intrinsic relationship between $\mathbf{c}_\text{p}$ and $\mathbf{z}_t$. Subsequently, we eliminate the part of image tokens and only reserve the part of noisy tokens for input to the next module. 

\noindent \textbf{Semantic-level alignment module.} We devise a semantic-level alignment module to ensure semantic consistency between the generated 3D models and the conditional images. We employ the pre-trained CLIP~\cite{radford2021clip} visual model (ViT-L/14) to extract semantic image tokens $\mathbf{c}_\text{s}$ from the conditional images, and then utilize a cross-attention layer within each DiT block to facilitate the interaction between $\mathbf{c}_\text{s}$ and noisy latent token $\mathbf{z}_t$. Meanwhile, unlike the original class conditional DiT, our image-conditioned diffusion model no longer utilizes class embedding. Instead, we use the classification token from the semantic image tokens $\mathbf{c}_\text{s}$ after projection and add it to the time embedding to enhance semantic features. In addition, to reduce the number of parameters and computational cost, we employ adaLN-single, as proposed in PixArt~\cite{chen2024pixartalpha}, which predicts a set of global shift and scale parameters $P=[\gamma_1, \beta_1, \alpha_1, \gamma_2, \beta_2, \alpha_2]$ using time embeddings, then sets a trainable embedding and adds it to $P$ for adjustment in each block.

\noindent \textbf{Training.} Following LDM~\cite{rombach2022high}, our 3D latent diffusion transformer model predicts the noise $\epsilon$ of the noisy latent representation $\mathbf{z}_\text{t}$ at time \text{t}, conditioned on image $C$. When training the diffusion model, we randomly zero the conditional input $\mathbf{c}_\text{p}$ and $\mathbf{c}_\text{s}$ with a probability of $10\%$ to use classifier-free guidance~\cite{ho2022cfg} during inference, thereby improving the quality of conditional generation.

%% file: sections/experiment.tex
\section{Experiments}

\begin{figure*}[!t]
  \centering
  \includegraphics[width=1.0\linewidth]{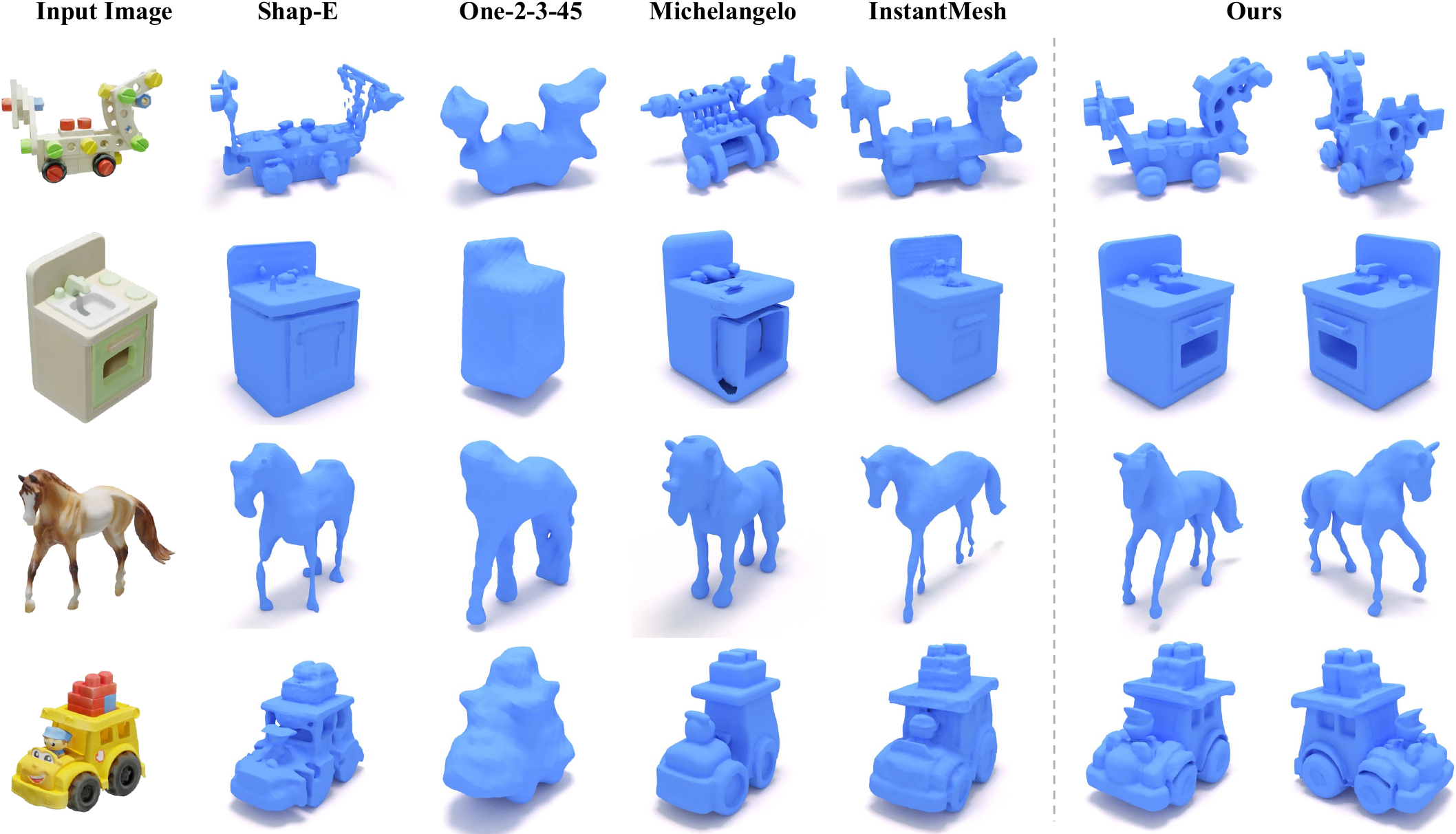}
  \caption{Qualitative comparisons with different baseline methods on GSO dataset.}
  \label{fig:compare_gso}
\end{figure*}

\begin{figure*}[!t]
  \centering
  \includegraphics[width=1.0\linewidth]{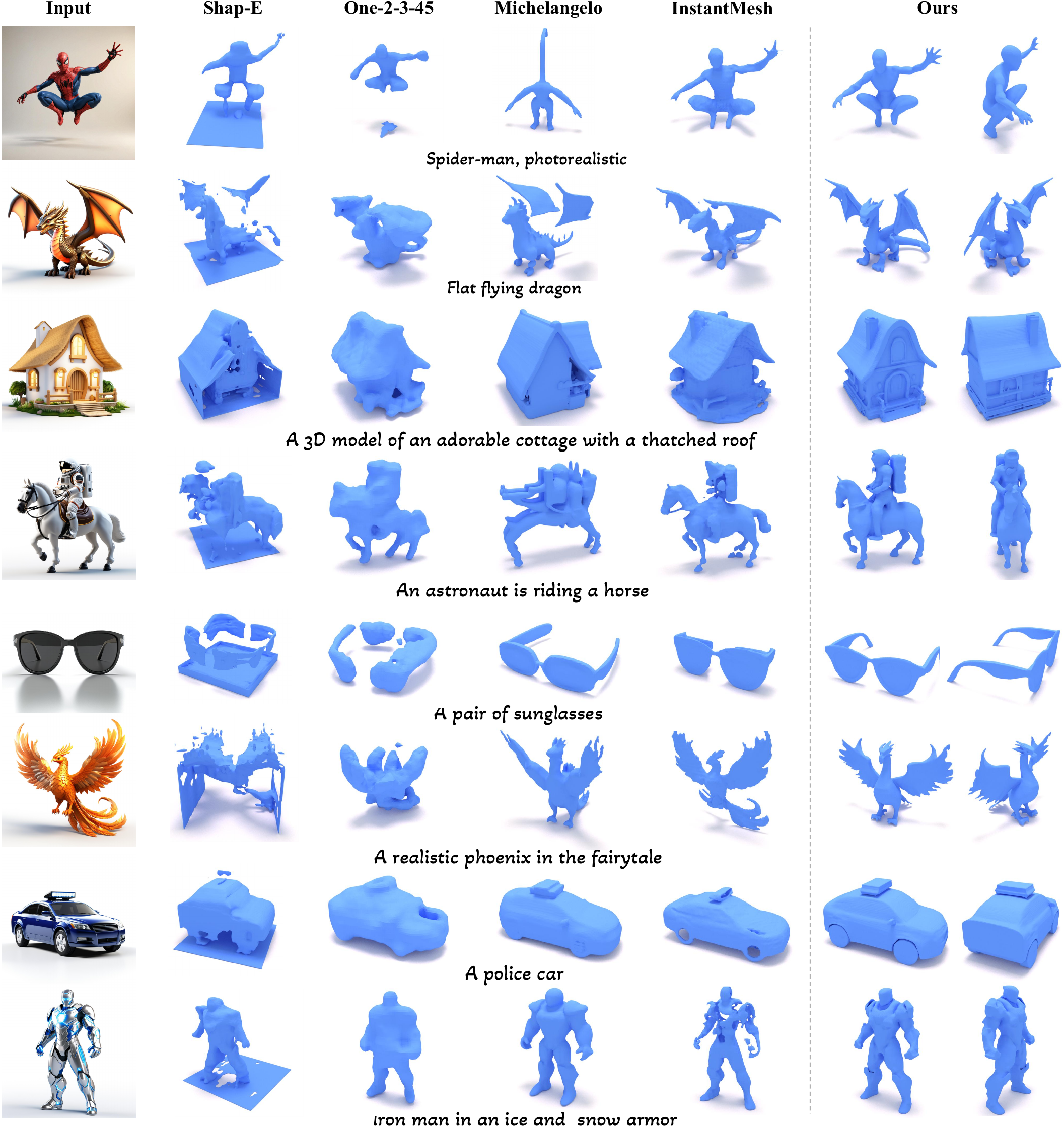}
  \caption{Qualitative comparisons of the meshes generated from text. We employ text-to-image diffusion model to produce highly detailed images as the inputs of each method.}
  \label{fig:compare_text}
\end{figure*}

\subsection{Implementation Details}
\noindent\textbf{D3D-VAE.}
Our D3D-VAE takes as input 81,920 point clouds with normal uniformly sampled from the 3D model, along with a learnable latent token of a resolution $r=32$ and a channel dimension $d_\mathbf{e}=768$. The encoder network consists of 1 cross-attention layer and 8 self-attention layers, with each attention layer comprising 12 heads of a dimension 64. The channel dimension of the latent representation is $d_\mathbf{z}=16$. The decoder network comprises of 5 ResNet~\cite{he2016resnet} blocks to upsample the latent representation into triplane feature maps with resolution of $256\times256$ and channel dimension of 32. The geometric mapping network consists of 5 linear layers with hidden dimension 64. During training, we sample 20,480 uniform points and 20,480 near-surface points for supervision. The KL regularization weight is set to $\lambda_\text{KL}=1e-6$. We use the AdamW~\cite{loshchilov2017adamw} optimizer with a learning rate $1e-4$. We train the D3D-VAE model on NVIDIA A100 (80G) for 100K steps.

\input{table/user_study}

\begin{figure*}[!t]
  \centering
  \includegraphics[width=1.0\linewidth]{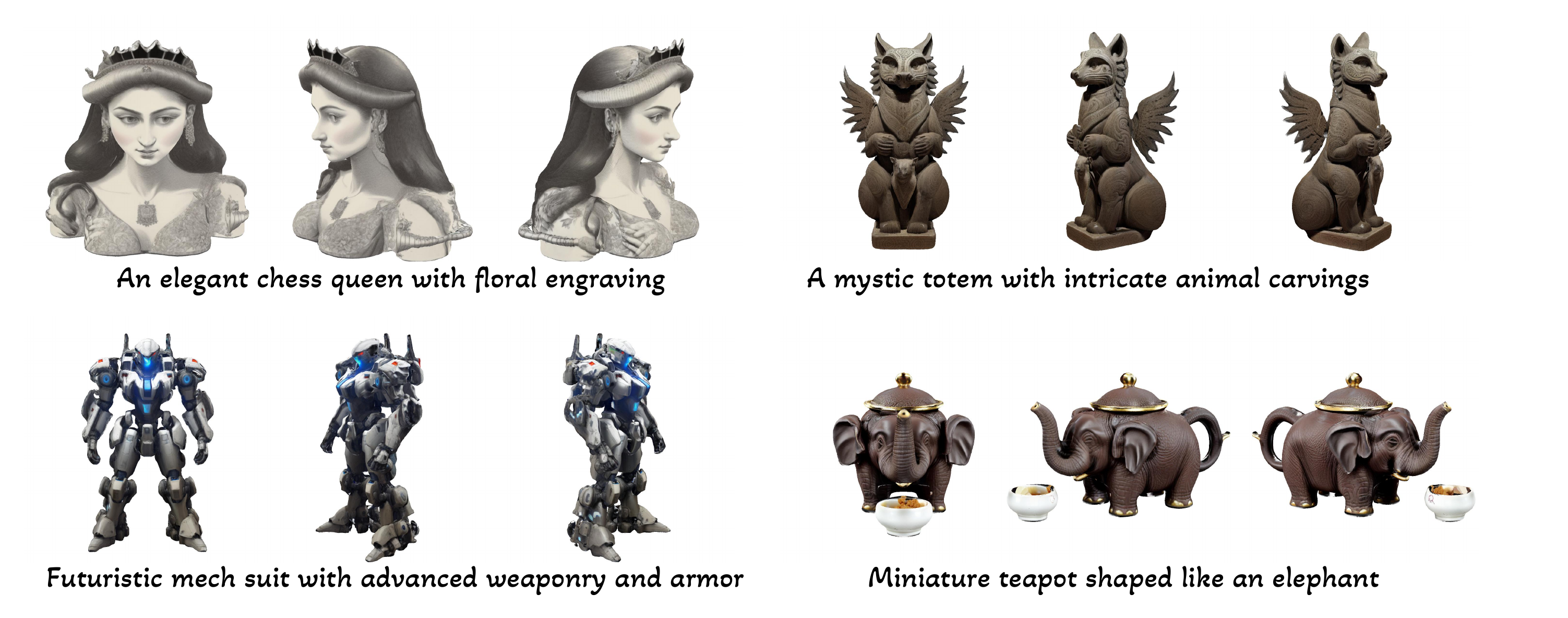}
  \caption{Visualizations of the textured meshes. We employ SyncMVD~\cite{liu2023syncmvd} to generate texture for the meshes produced by our Direct3D.}
  \label{fig:texture_vis}
\end{figure*}

\noindent\textbf{D3D-DiT.}
Our diffusion model adopts the network configuration of DiT-XL/2~\cite{peebles2023dit}, which consists of 28 layers of DiT blocks. Each attention layer includes 16 heads with a dimension of 72. We train the diffusion model with 1000 denoising steps using a linear variance scheduler ranging from $1e-4$ to $2e-2$. We employ the AdamW optimizer with a learning rate $1e-4$ and train for 800K steps. During inference, we apply 50 steps of DDIM~\cite{song2020ddim} with the guidance scale set to 7.5.


\subsection{Image and Text to 3D Generation}
\noindent\textbf{Image-to-3D.}
We conduct qualitative comparisons of our Direct3D with other baseline methods on the GSO dataset for the image-to-3D task, as illustrated in Figure~\ref{fig:compare_gso}. Shap-E~\cite{jun2023shapee}, a 3D diffusion model trained on millions of 3D assets, is capable of producing plausible geometry, but it suffers from artifacts and holes in the meshes. Michelangelo~\cite{zhao2024michelangelo} performs diffusion process on a 1D implicit latent space, and fails to align the generated mesh with the semantic content of the conditional images. Multi-view based approaches such as One-2-3-45~\cite{liu2023one2345} and InstantMesh~\cite{xu2024instantmesh} heavily rely on the performance of multi-view 2D diffusion model. One-2-3-45 directly employs SparseNeuS~\cite{long2022sparseneus} for reconstruction, resulting in coarse geometry. The meshes generated by InstantMesh perform decent quality, but lack consistency with the input images in certain details like the water spout on the sink and the windows of the school bus. It also produces some failure cases such as merging the hind legs of the horse together, due to the limitation of multi-view diffusion model. In contrast, our Direct3D consistently generates high-quality meshes that align with the conditional images in most cases.

\noindent\textbf{Text-to-3D.}
Our Direct3D can produce 3D assets from text prompts by incorporating text-to-image models like Hunyuan-DiT~\cite{li2024hunyuan}. Figure~\ref{fig:compare_text} illustrates the qualitative comparisons of our Direct3D and other baseline methods on the text-to-3D task. To ensure a fair comparison, all methods utilize the same generated image as input. It can be observed that these baseline methods fail in almost all cases, while our Direct3D is still able to generate high-quality meshes, demonstrating the generalizability of our approach. We also conducted a user study to quantitatively compare our D3D-DiT with other methods. We render videos of meshes generated by each method rotating 360 degrees, and ask 46 volunteers to rate each mesh based on its quality and consistency with the input images. The results in Table~\ref{tab:user_study} indicate that our D3D-DiT perform superior mesh quality and consistency compared to other baseline methods.

\noindent\textbf{Generation of textured mesh.} Benefited from the smooth and detailed geometry produced by our Direct3D, we can easily dress up the mesh using existing texture synthesis methods. As shown in Figure~\ref{fig:texture_vis}, we utilize SyncMVD~\cite{liu2023syncmvd} to obtain exquisite textured meshes.






%% file: table/user_study.tex
\begin{table}[!t]
  \centering
  \caption{User study on the quality of meshes. The higher the score, ranging from 1 to 5, the better.}
    \small
  \begin{tabular}{l | c | c | c | c | c}
    \toprule
      & Shap-E~\cite{jun2023shapee} & One-2-3-45~\cite{liu2023one2345} & Michelangelo~\cite{zhao2024michelangelo} & InstantMesh~\cite{xu2024instantmesh} & Ours\\
    \midrule
    Quality&1.18&1.24&2.51&2.53&\textbf{4.41} \\
    Consistency &1.19&1.28&2.32&2.66&\textbf{4.35} \\
    \bottomrule
  \end{tabular}
  \label{tab:user_study}
\end{table}

%% file: sections/conclusion.tex
\section{Conclusion}

In conclusion, our paper introduces a novel approach for direct 3D shape generation from a single image, bypassing the need for multi-view reconstruction. Leveraging a hybrid architecture, our proposed D3D-VAE efficiently encode 3D shapes into a compact latent space, enhancing the fidelity of the generated shapes. Our image-conditioned 3D diffusion transformer (D3D-DiT) further improves the generation quality by integrating image information at both pixel and semantic levels, ensuring high consistency between generated 3D shapes and conditional images. Extensive experiments on the image-to-3D and text-to-3D tasks demonstrate the superior performance of our Direct3D in 3D generation, surpassing existing methods in quality and generalizability. 

\noindent\textbf{Limitations.} Despite the capability of our Direct3D to produce high-fidelity 3D assets, it is currently limited to the generation of individual or multiple objects and cannot generate large-scale scenes. We will focus on it in future research.

%% file: sections/appendix.tex
\section{Appendix}

\subsection{Ablation Studies}

\begin{figure*}[!h]
  \centering
  \includegraphics[width=1.0\linewidth]{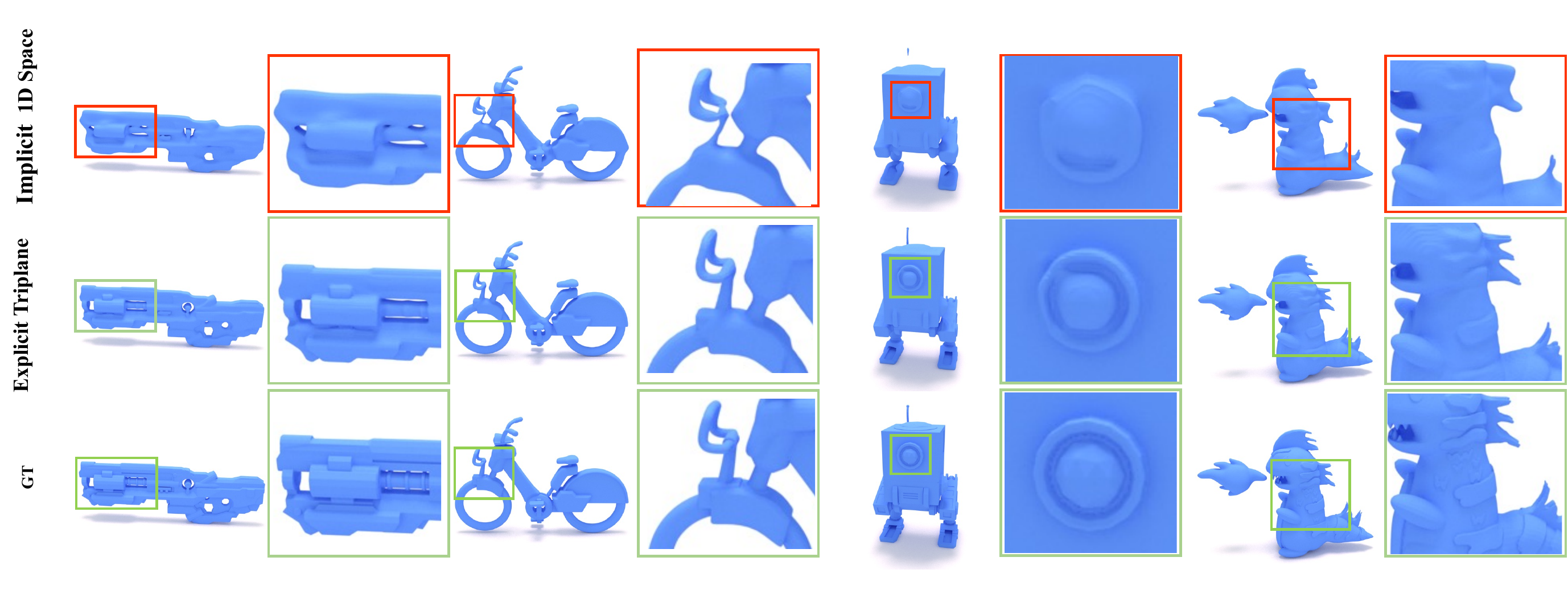}
  \caption{Qualitative comparisons of reconstruction with different latent representation.}
  \label{fig:vae_latent}
\end{figure*}

\noindent \textbf{Explicit triplane latent.} Unlike typical approaches like Michelangelo~\cite{zhao2024michelangelo} which employs VAE to encode inputs into a 1D implicit latent space, our D3D-VAE compresses high-resolution point clouds into an explicit triplane latent representation.
Figure~\ref{fig:vae_latent} illustrates the comparison of the reconstruction results of VAE between these two approaches, demonstrating that our adopted explicit triplane representation is more capable of recovering high-frequency geometric details.

\begin{figure*}[!t]
  \centering
  \includegraphics[width=1.0\linewidth]{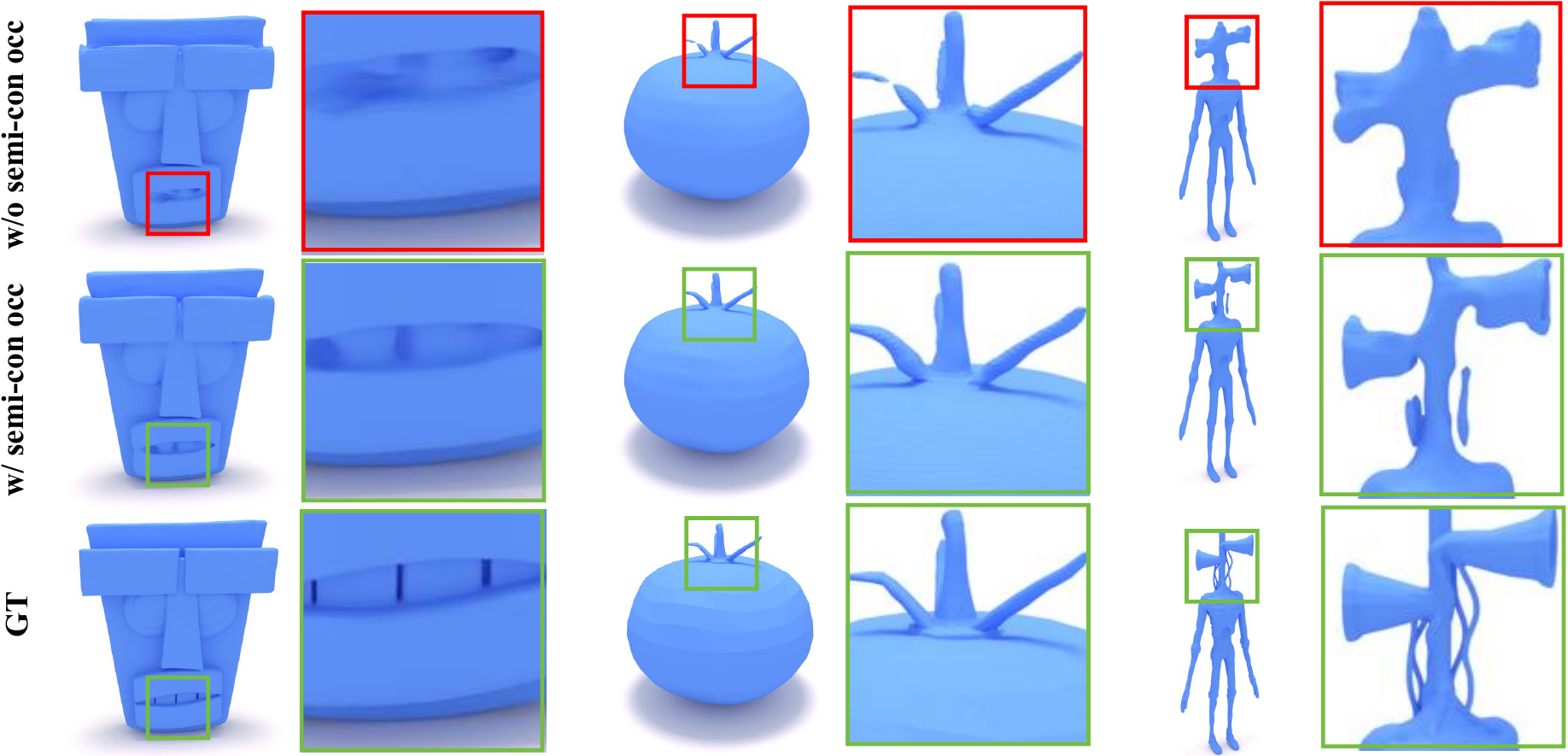}
  \caption{Ablation study for the semi-continuous surface sampling strategy.}
  \label{fig:semi}
\end{figure*}

\noindent \textbf{Semi-continuous surface sampling strategy.} We adopt a semi-continuous surface sampling strategy during the training of D3D-VAE to alleviate the optimization difficulty caused by abrupt changes in occupancy near the object surface. To evaluate the effectiveness of this strategy, we train D3D-VAE with and without this sampling strategy separately and compare the reconstructed results. As shown in Figure~\ref{fig:semi}, it can be observed that the reconstruction performance is unsatisfactory when directly training with the original occupancy in some thin structures, but is improved when the semi-continuous sampling strategy is utilized.

\begin{figure*}[!t]
  \centering
  \includegraphics[width=1.0\linewidth]{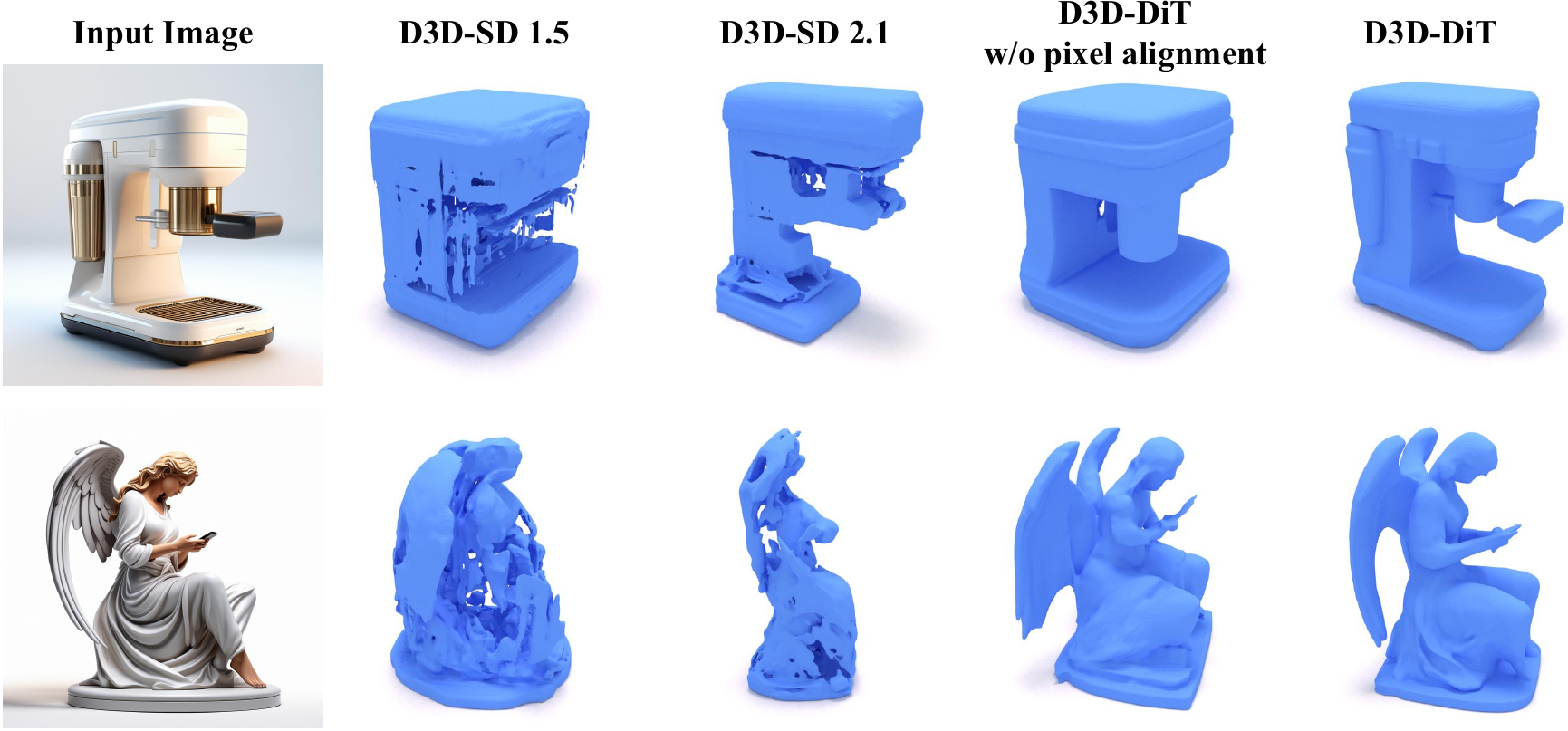}
  \caption{Qualitative comparisons of diffusion models with different network architectures.}
  \label{fig:diffusion}
\end{figure*}

\noindent \textbf{2D U-Net vs D3D-DiT.} To demonstrate the superiority of our D3D-DiT network architecture, we conduct experiments to compare it with 2D U-Net. We train diffusion models on the roll-out triplane latent representation using network architectures of SD 1.5~\cite{rombach2022high} and SD 2.1, respectively. Figure~\ref{fig:diffusion} illustrates the qualitative comparisons using the conditional images generated by Hunyuan-DiT~\cite{li2024hunyuan}. It can be observed that neither SD 1.5 or SD 2.1 is able to produce satisfactory meshes, while our D3D-DiT, due to its powerful scalability and generalization, is capable of generating high-quality 3D shapes that align with the content of the conditional images.

\begin{figure*}[!t]
  \centering
  \includegraphics[width=1.0\linewidth]{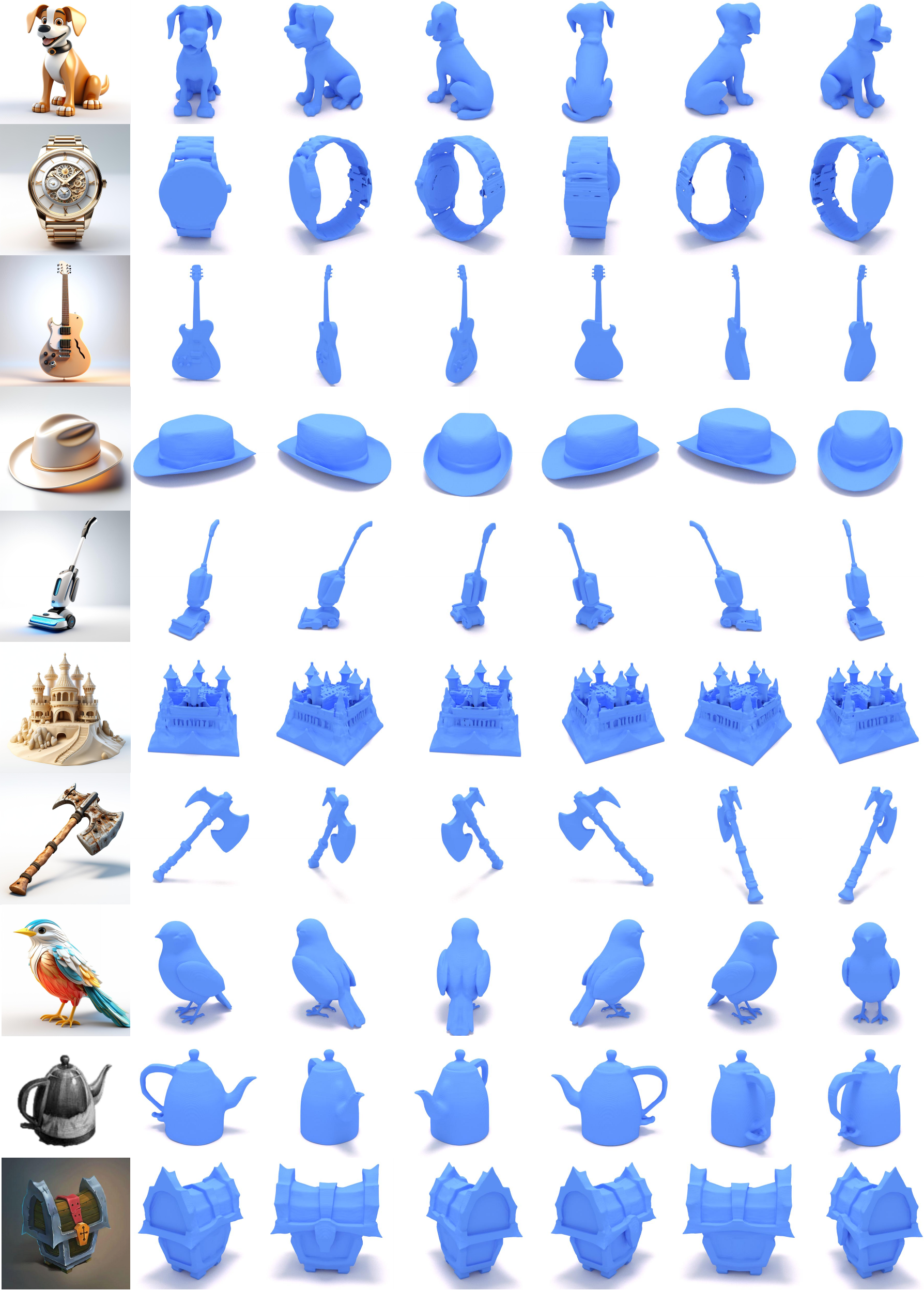}
  \caption{More Visualizations.}
  \label{fig:more_vis_0}
\end{figure*}

\noindent \textbf{Effectiveness of the pixel-level alignment module.} We perform ablation experiments to validate the effectiveness of the pixel-level alignment module used in our D3D-DiT. As illustrated in Figure~\ref{fig:diffusion}, D3D-DiT can still generate meshes of relatively high quality without this module. However, it does not align well with the conditional images, such as the external structure of the coffee machine and the wings of the statue. By injecting the pixel-level information into each DiT block through the pixel-level alignment module, the produced meshes can also maintain consistency with the conditional images in terms of details.

\subsection{More visualizations}
We present more visualizations in Figure~\ref{fig:more_vis_0}.